\documentclass[runningheads]{llncs}

\usepackage[mobile]{eccv}

\usepackage{eccvabbrv}

\usepackage{graphicx}
\usepackage{booktabs}
\usepackage{multirow}
\usepackage{gensymb}

\usepackage[accsupp]{axessibility}  

\usepackage{colortbl}
\usepackage{wrapfig}
\usepackage{caption}
\definecolor{first}{rgb}{1.0, .83, 0.3}
\definecolor{second}{rgb}{1.0, 0.93, 0.7}
\def \first {\cellcolor{first}}
\def \second {\cellcolor{second}}

\usepackage{hyperref}

\usepackage{orcidlink}

\begin{document}

\title{FixAnything: 3D-Consistent Rendering Refinement via Video Generative Priors} 

\titlerunning{FixAnything}

\author{Khiem Vuong \and
Deva Ramanan$^{\ast}$ \and
Srinivasa Narasimhan$^{\ast}$}

\authorrunning{K.~Vuong et al.}

\newcommand{\MYhref}[3][0000AA]{\href{#2}{\color[HTML]{#1}{\textbf{#3}}}}%

\institute{Carnegie Mellon University, USA\\
{\normalsize \MYhref{https://fix-anything.github.io}{\texttt{https://fix-anything.github.io}}}
}

\maketitle

\vspace{-0.3in}

\begin{figure}
    \centering
    \includegraphics[width=0.90\linewidth]{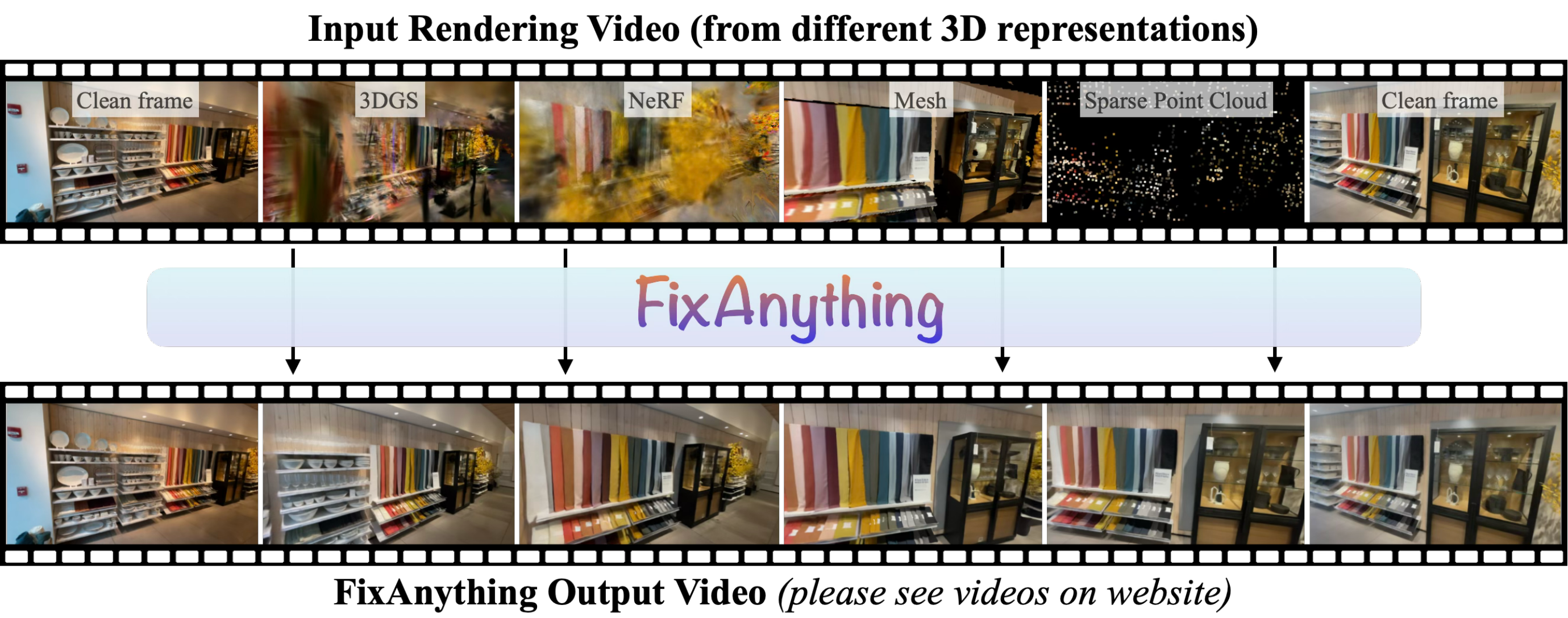}
    \vspace{-0.10in}
    \captionsetup{width=0.90\linewidth}
    \caption{\textbf{FixAnything} takes a rendering video from any 3D representation --- 3DGS, NeRF, mesh, or sparse point cloud --- and produces a photorealistic, 3D-consistent video that preserves the camera trajectory and scene content, using a single model finetuned from a pretrained video diffusion model with minimal adaptation. \emph{Top:} One frame selected from each of the four videos rendered using different 3D representations. Clean frames are the input training views (they do not need to be at the start/end). \emph{Bottom:} Corresponding clean output frames showing quality, consistency \& generality.}
    \label{fig:teaser}
\end{figure}
\vspace{-0.4in}

{\renewcommand\thefootnote{}\NoHyper\footnote{$^{\ast}$ denotes equal contribution/advising}\endNoHyper}

\begin{abstract}
\vspace{-0.15in}
Rendering views using 3D scene representations such as Gaussian Splatting (3DGS), Neural Radiance Fields (NeRF), meshes, or even point clouds produces artifacts when input views are sparse or target views lie far from the input.
Recent work mitigates these artifacts using diffusion-based generative priors, but is specialized to individual representations and require custom architectures or extensive retraining.
We present FixAnything, a single model for fixing a wide range of rendering artifacts. It does so by repurposing a pretrained video generative model, leveraging its implicit multi-view priors with only minimal modification and lightweight finetuning.
%
%
Our key insight is that even noisily-rendered sequences preserve camera motion and coarse scene structure, allowing cleanup to be formulated as video-to-video translation.
To control what scene structure should be preserved, we introduce a binary mask denoting the clean pixels, enabling the model to anchor its output to high-quality inputs (e.g. training views) while refining the rest.
%
%
To encourage FixAnything to produce 3D-consistent renderings that support downstream reconstruction, we use camera pose accuracy (recovered via structure-from-motion) as a reward signal for direct preference optimization (DPO).
Across four distinct 3D representations, FixAnything consistently improves rendering quality with lightweight finetuning, demonstrating that a single generalist video prior can replace multiple specialist refinement pipelines.
The simplicity of the framework enables immediate adoption of stronger future video models without architectural redesign.

\end{abstract}

\section{Introduction}
\label{sec:intro}


The quality of 3D reconstruction has improved dramatically in recent years, but a gap remains: when input views are sparse or novel viewpoints lie far from training views, every 3D representation produces artifacts.
3DGS~\cite{kerbl3Dgaussians} produces floaters across the scene, NeRF~\cite{mildenhall2020nerf} hallucinates foggy geometry, mesh reconstructions suffer from texture distortions, and point clouds leave holes. These artifacts directly limit downstream applications, where the visual quality of novel views may degrade to the point where they are unusable for content creation or robotics.
The dominant approach to improve the quality of these renderings has been to build a specialist generative pipeline that removes the artifacts of that particular representation: 3D or per-image diffusion priors for NeRF~\cite{warburg2023nerfbusters, wu2025difix3d+, wu2024reconfusion}, video diffusion models tailored to 3DGS~\cite{liu20243dgs, xu2025novel}, and camera-controlled video generation conditioned on explicit geometry~\cite{yu2024viewcrafter, ren2025gen3c, gu2025diffusion}.
While effective individually, such a bespoke approach is difficult to scale: each time a new representation emerges or an existing one is improved, a new pipeline must be built, with its own architecture, training data, and conditioning mechanism.


This raises a natural question: can a single generalist replace this growing family of specialists? We show the answer is yes. Despite looking visually distinct, artifacts from different representations share a common property: they all deviate from the manifold of natural videos, while preserving the underlying camera trajectory and coarse scene layout. A pretrained video model, with minimal adaptation, can exploit this shared structure to translate (or project) degraded renderings onto the manifold of natural videos.

We present \textbf{FixAnything}, a framework that realizes this generalist approach.
FixAnything takes as input a video rendered along a camera trajectory and directly produces a cleaned version, naturally preserving temporal coherence across views (see \cref{fig:teaser}).
We make use of a video diffusion model (Wan2.1~\cite{wan2025wan}) with minor architectural changes: the rendering video is concatenated as conditioning input in  latent space, and only a lightweight LoRA~\cite{hu2022lora} is trained to adapt the model for this task.
A per-frame binary mask distinguishes clean reference frames (which the model should trust and preserve) from degraded frames (which need fixing), anchoring each restoration to known high-quality views (e.g. training views).
We generate training pairs from DL3DV-10K~\cite{ling2024dl3dv} using videos rendered from four different 3D representations including 3DGS, NeRF, meshes, and sparse point clouds. Because the pretrained model already understands natural videos, as few as 20 paired videos suffice for effective training.

Interestingly, we find that different 3D representations achieve comparable quality after cleanup. This is particularly remarkable for sparse (COLMAP~\cite{schonberger2016structure}) point clouds, which often act as a prerequisite for other 3D representations such as 3DGS and NeRF (since they require {\em posed} input images). Our results suggest that such ``intermediate'' 3D representations may not be needed given a sufficiently powerful generative model. Indeed, video generation can already interpolate between two input frames (e.g., ``first-last-frame-to-video'' generation~\cite{wan2025wan}). However, such models generate the most-likely camera path rather than conditioning on one provided by a rendering engine. We show that sparse point cloud renders are sufficient for exposing this camera path, avoiding the need to teach the generative model about $\mathrm{SE}(3)$ camera coordinates (e.g., using Pl\"ucker coordinates~\cite{plucker1865xvii, zhang2024raydiffusion} which typically requires far more than 20 finetuning videos~\cite{zhu2025aether}).


However, one final challenge is that the generated videos may not be 3D consistent, which may prevent downstream applications such as 3D reconstruction. Rather than building geometric constraints into the architecture, we treat geometric consistency as a preference optimization problem: for each training video, we sample multiple outputs with different random seeds, rank them by how accurately structure-from-motion~\cite{schonberger2016structure} recovers their camera poses, and apply Flow-DPO~\cite{liu2025improving} to steer the model toward geometrically coherent outputs.
This bakes the geometric consistency into the model during training, improving pose estimation accuracy by 7.2\% (AUC@5\degree{}) with no additional inference cost.

Despite its simplicity, FixAnything matches or exceeds specialist methods while generalizing across representations.
Because the adaptation requires no architectural changes and updates less than 1\% of parameters, the same recipe can be applied to newer video foundation models as they become available, requiring only a new LoRA training run on limited (academic-scale) compute.

In summary, our contributions are:
\begin{itemize}
    \item A generalist framework that takes rendering videos along camera trajectories of a 3D representation as input and fixes artifacts with a single model, without architectural changes to the base video model.
    \item A mask-aware conditioning mechanism that distinguishes \textit{frames to trust} from \textit{frames to fix}, combined with data-efficient training that achieves effective results with limited paired data.
    \item A geometry-aware preference optimization that bakes multi-view consistency into the model using camera pose accuracy as a reward signal via Flow-DPO.
\end{itemize}

\section{Related Work}
\label{sec:related}

\paragraph{Sparse-view novel view synthesis.}
NeRF~\cite{mildenhall2020nerf} and 3DGS~\cite{kerbl3Dgaussians} achieve high-fidelity rendering when dense input views are available, but their quality degrades sharply under sparse-view settings due to the lack of multi-view supervision.
Many methods address this by introducing regularization during reconstruction.
RegNeRF~\cite{Niemeyer2021Regnerf} penalizes rendered patches at unobserved viewpoints to smooth geometry, while FreeNeRF~\cite{Yang2023FreeNeRF} applies frequency and occlusion regularization to prevent overfitting.
DSNeRF~\cite{Deng_2022_CVPR} and SparseNeRF~\cite{wang2023sparsenerf} leverage depth cues to stabilize geometry under sparse supervision.
On the 3DGS side, DNGaussian~\cite{li2024dngaussian} and FSGS~\cite{zhu2024fsgs} also introduce similar geometric constraints to improve stability in sparse-view settings.
While these approaches reduce artifacts, they remain constrained by the available observations. This faithfulness to input views is desirable when sufficient multi-view coverage exists, but it also limits their ability to infer plausible content in regions that are weakly observed or unseen. In such cases, incorporating generative priors can provide reasonable completions while maintaining consistency with observed views.

\paragraph{Generative priors for novel view synthesis and 3D enhancement.}
Recent work leverages pretrained generative models to improve 3D reconstructions beyond what the input views alone can support.
Nerfbusters~\cite{warburg2023nerfbusters} trains a 3D diffusion prior to directly regularize NeRF geometry, removing floater artifacts in 3D space.
ReconFusion~\cite{wu2024reconfusion} generates novel views one at a time, conditioned on PixelNeRF~\cite{yu2021pixelnerf} features from nearby cameras, and uses them to regularize NeRF training.
Difix3D+~\cite{wu2025difix3d+} applies a single-step image diffusion model conditioned on a reference view to fix individual renderings, then progressively distills the cleaned images back into the 3D representation.
FlowR~\cite{fischer2025flowr} trains a multi-view flow matching model that jointly processes multiple views to map degraded renderings from sparse reconstructions to their dense-reconstruction counterparts.
These methods process views independently or in small multi-view sets, and rely on the underlying 3D representation to enforce global consistency.

Other works repurpose \textit{video diffusion models}~\cite{blattmann2023stable, wan2025wan, yang2024cogvideox, agarwal2025cosmos} for 3D and novel view synthesis tasks.
ViewCrafter~\cite{yu2024viewcrafter} and GEN3C~\cite{ren2025gen3c} use camera-controlled video generation conditioned on rendered point clouds to synthesize novel views from sparse inputs.
3DGS-Enhancer~\cite{liu20243dgs} trains a video latent diffusion model with a custom spatial-temporal decoder to restore view-consistent renderings, then finetunes the 3DGS model on the enhanced views.
Xu~et~al.~\cite{xu2025novel} recast sparse-view NVS as test-time video completion, using a pretrained video diffusion model with uncertainty-aware modulation to generate pseudo-views that densify 3DGS supervision.
A common pattern across all these methods is specialization: each targets a specific representation, introduces custom architecture components, and requires large-scale paired or annotated data.
FixAnything instead adapts a single pretrained video model to handle four representation types with no architectural changes and orders of magnitude less training data.

\paragraph{Preference optimization for diffusion models.}
Direct Preference Optimization (DPO)~\cite{rafailov2023direct} was originally proposed for aligning language models with human preferences without training an explicit reward model.
The idea has since been extended to generative vision models. For instance, Diffusion-DPO~\cite{wallace2024diffusion} adapts the framework to noise-prediction diffusion models for image generation.
For modern video generators based on rectified flow~\cite{lipman2022flow, liu2022flow}, Flow-DPO~\cite{liu2025improving} reformulates the preference loss in terms of velocity prediction and demonstrates improvements in visual quality and text alignment for text-to-video models.
These prior works optimize for human aesthetic preferences or prompt fidelity.
As concurrent works, Epipolar-DPO~\cite{kupyn2025epipolar} uses the Sampson distance from epipolar geometry as a reward signal to improve 3D consistency in text-to-video and image-to-video generation, and VideoGPA~\cite{du2026videogpa} distills dense geometric priors from reconstruction foundation models into video diffusion models via DPO.
Our work shares the motivation of using geometric consistency as a preference signal but applies it to a different setting: rather than improving general-purpose video generation from text or a single image, we target rendering cleanup for 3D reconstructions, using camera pose accuracy from structure-from-motion as the reward to ensure that refined videos support downstream 3D tasks.

\section{Method}
\label{sec:method}


\begin{figure}[t]
    \centering
    \includegraphics[width=\linewidth]{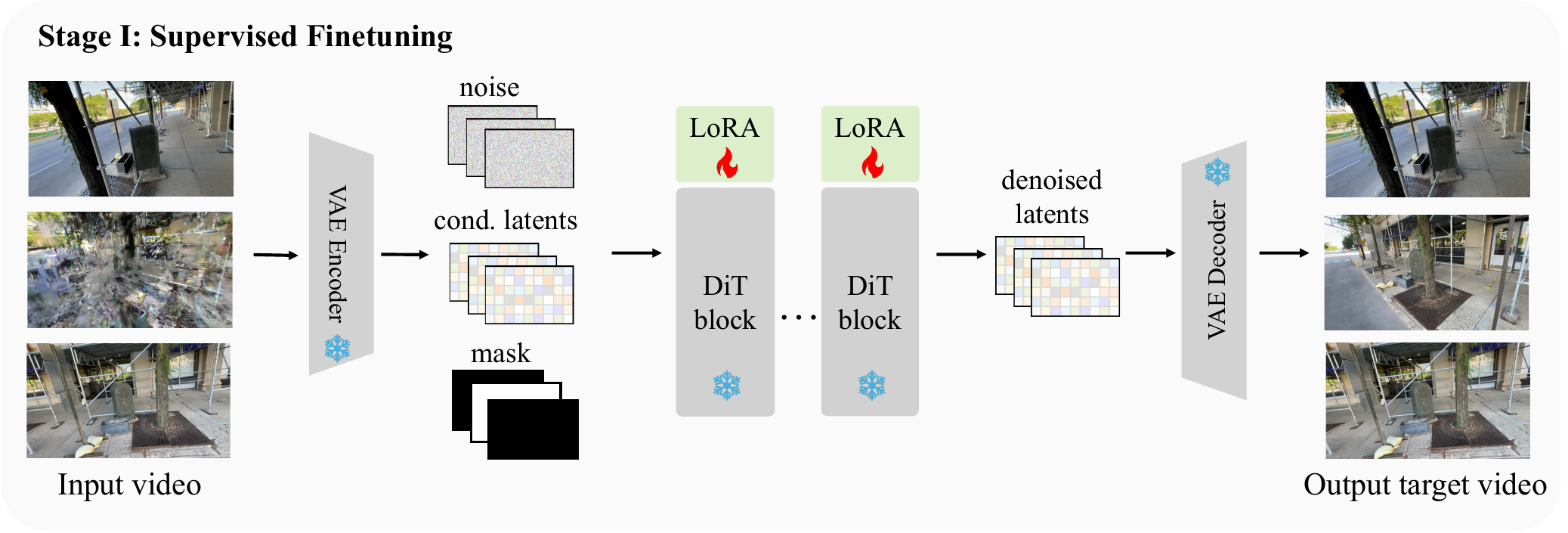}\\[4pt]
    \includegraphics[width=\linewidth]{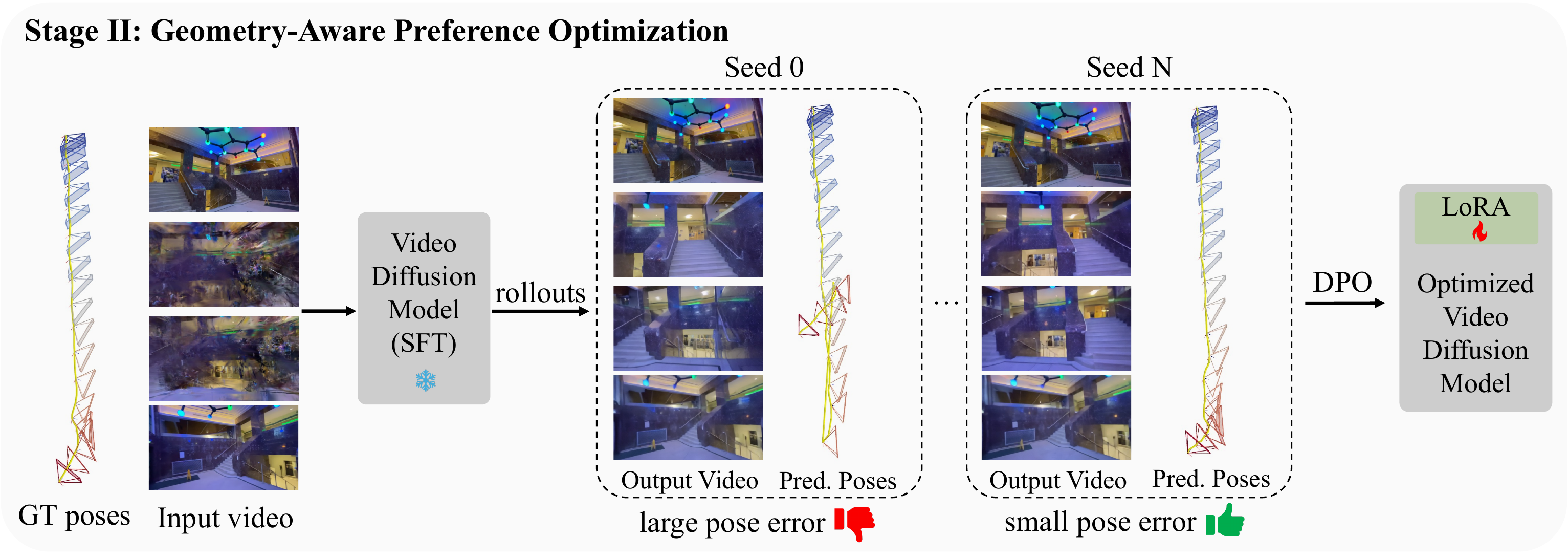}
    \vspace{-0.25in}
    \caption{\textbf{Overview of FixAnything.}
    Given a rendered video from any 3D representation (3DGS, NeRF, mesh, sparse point cloud) and a binary mask indicating clean frames (e.g., training views), FixAnything produces a cleaned video that preserves scene content and camera trajectory.
    \textit{Top:} the rendered video is encoded into latent space via a frozen VAE, which is channel-concatenated with the binary mask and noise. These inputs are denoised with a Wan2.1 DiT, adapted only through LoRA~\cite{hu2022lora}, without any architectural modification.
    \textit{Bottom:} after supervised finetuning (SFT), we further optimize the model by generating multiple candidate outputs per video with different random seeds and evaluate their geometric consistency by using COLMAP to recover camera poses.
    The resulting pose accuracy is used to construct preference pairs for Flow-DPO, which steers the model toward outputs that are geometrically consistent.}
    \label{fig:pipeline}
\end{figure}

\cref{fig:pipeline} provides an overview of FixAnything.
We first formulate the problem as representation-agnostic rendering cleanup (\cref{sec:problem}), then describe how a pretrained video diffusion model is adapted for this task with minimal changes (\cref{sec:adaptation}).
\cref{sec:training_data} details the training data, and \cref{sec:geometry_dpo} introduces a preference optimization stage that further encourages geometric consistency for the model.

\subsection{Representation-Agnostic Rendering Cleanup}
\label{sec:problem}

Given any 3D representation, we can render a video along an arbitrary camera trajectory.
Some frames along this trajectory, at exact or near training viewpoints, will look clean, while others contain artifacts.
Rather than fixing frames independently, FixAnything processes the entire rendering video at once, leveraging temporal context from  clean frames to guide the cleanup of degraded ones.

\begin{figure*}[t!]
    \centering
    \includegraphics[width=0.95\linewidth]{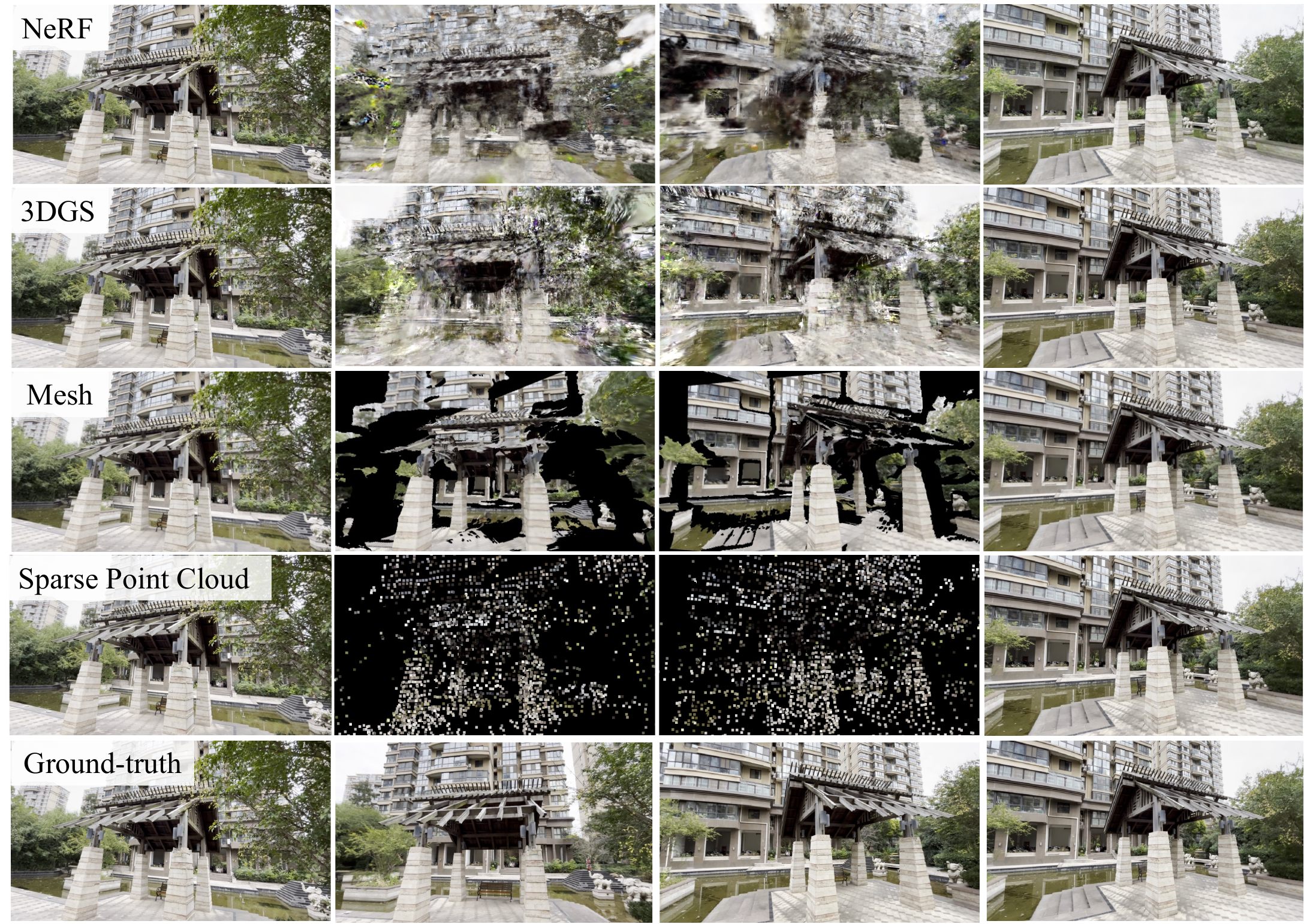}
    \vspace{-0.1in}
    \caption{\textbf{Different types of degraded renderings} from the same scene, paired with ground truth (last row).
    First and last columns are clean training views, while intermediate frames show artifacts of varying severity.
    NeRF produces blur and fog (row~1), 3DGS produces floaters (row~2), meshes have holes at ambiguous depth regions (row~3), and sparse point clouds render as scattered patches at detected keypoints (row~4).
    Despite their visual differences, all representations preserve the underlying camera trajectory and coarse scene layout, and FixAnything trains on all four types jointly.}
    \label{fig:artifact_types}
\end{figure*}

Concretely, let $\mathbf{x} \in \mathbb{R}^{T \times 3 \times H \times W}$ be a rendering video along a camera trajectory, and let $\mathbf{m} \in \{0, 1\}^{T}$ be a binary mask where $\mathbf{m}_i = 1$ marks frames rendered from training viewpoints (clean) and $\mathbf{m}_i = 0$ marks the rest (degraded).
FixAnything produces a clean video $\mathbf{y} \in \mathbb{R}^{T \times 3 \times H \times W}$ that preserves scene content and camera motion while fixing degraded frames.
This formulation is representation-agnostic as renderings from NeRF, 3DGS, meshes, and point clouds look different, but all preserve the camera trajectory and coarse scene layout, providing the structure a pretrained video model can leverage to remove artifacts (\cref{fig:artifact_types}).


\subsection{Lightweight Adaptation of a Pretrained Video Model}
\label{sec:adaptation}

FixAnything builds on Wan2.1-I2V-14B~\cite{wan2025wan}, a DiT-based~\cite{peebles2023scalable} image-to-video diffusion model trained with rectified flow.
We repurpose it for rendering cleanup by channel-wise concatenating the rendering video as conditioning signal and training a lightweight LoRA~\cite{hu2022lora} adapter, keeping the architecture unchanged.

\paragraph{Conditioning via channel concatenation.}
We operate in the latent space of the pretrained VAE.
Let $\mathbf{z}_{\mathrm{cond}} = \mathcal{E}(\mathbf{x})$ denote the VAE-encoded latent of the degraded video and $\mathbf{z}_0 = \mathcal{E}(\mathbf{y})$ the latent of the clean target.
At each timestep $t \in [0, 1]$, we form the noised latent via the rectified flow interpolation:
\begin{equation}
    \mathbf{z}_t = (1 - t)\,\mathbf{z}_0 + t\,\boldsymbol{\epsilon}, \quad \boldsymbol{\epsilon} \sim \mathcal{N}(\mathbf{0}, \mathbf{I}).
    \label{eq:interpolation}
\end{equation}
The degraded video latent $\mathbf{z}_{\mathrm{cond}}$ carries the camera trajectory and coarse scene layout that the model should preserve; we inject it by concatenating with $\mathbf{z}_t$ and a spatially broadcast version of the mask $\mathbf{m}$ along the channel dimension:
\begin{equation}
    \hat{\mathbf{z}}_t = [\mathbf{z}_t \,;\, \mathbf{z}_{\mathrm{cond}} \,;\, \mathbf{m}],
    \label{eq:concat}
\end{equation}
where $[\cdot\,;\,\cdot]$ denotes channel-wise concatenation.
Given this augmented input, the model $\mathbf{v}_\theta$ predicts the velocity field $\mathbf{v} = \boldsymbol{\epsilon} - \mathbf{z}_0$ and is trained with the flow matching objective~\cite{liu2022flow, lipman2022flow}:
\begin{equation}
    \mathcal{L}_{\mathrm{FM}} = \mathbb{E}_{\mathbf{z}_0,\, \boldsymbol{\epsilon},\, t}\left[\left\| \mathbf{v} - \mathbf{v}_\theta(\hat{\mathbf{z}}_t, t) \right\|^2\right].
    \label{eq:flow_matching}
\end{equation}

\paragraph{Mask-aware conditioning: what to trust vs.\ what to fix.}
A rendering video from a sparse reconstruction is not uniformly degraded.
Frames near training viewpoints may look nearly perfect, while frames further away can be severely corrupted.
Naively asking a generative model to ``clean up'' the entire video has a problem: the model cannot easily distinguish frames that are already correct from those that need fixing, and may hallucinate over content that should be preserved.

\begin{figure*}[t!]
    \centering
    \includegraphics[width=\linewidth]{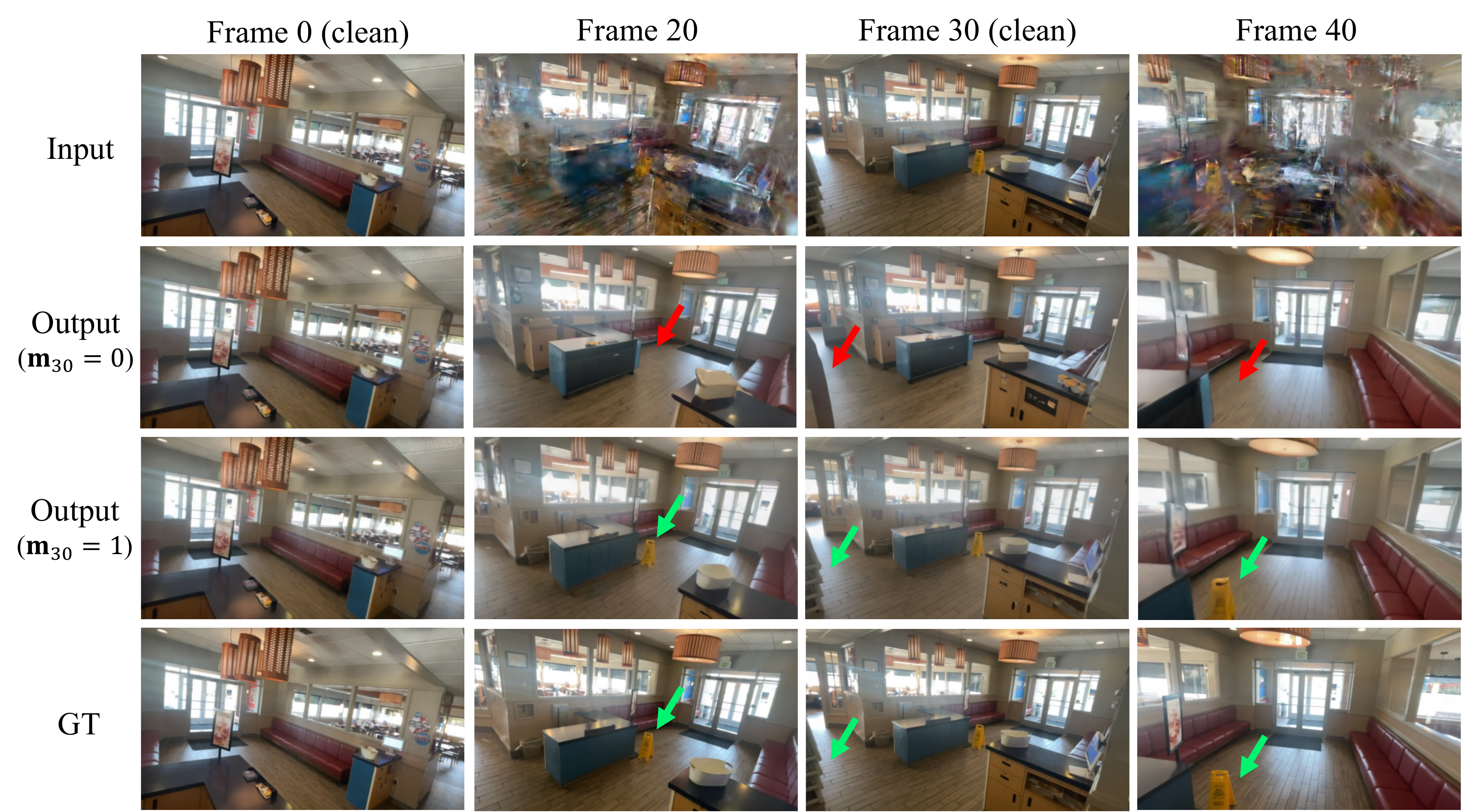}
    \vspace{-0.25in}
    \caption{\textbf{Effect of mask-aware conditioning in cleaning up 3DGS inputs}.
    The input video (row~1) has clean frames at positions 0 and 30.
    Without the mask (row~2, $\mathbf{m}_{30}{=}0$), the model treats frame~30 as degraded and hallucinates over it: objects like the wet floor sign disappear (red arrows).
    This corruption propagates to nearby frames, causing frame~40 to lose content that should be preserved.
    With the mask (row~3, $\mathbf{m}_{30}{=}1$), the model preserves frame~30 and uses it as context for neighboring ones: the wet floor sign and details are retained (green arrows), matching ground truth (row~4).}
    \label{fig:mask_comparison}
\end{figure*}

The mask $\mathbf{m}$ resolves this by making the distinction explicit: entries corresponding to training poses are set to $1$ (trust) and all others to $0$ (fix).
This provides two signals.
First, it knows which frames to leave untouched, preventing unnecessary hallucination over already-clean content.
Second, the clean frames become anchors that provide context for the degraded ones: the model can propagate appearance, lighting, and scene structure from trusted/clean frames to their neighbors, rather than guessing from scratch.
Although the mask itself is binary, its temporal arrangement is informative: degradation severity typically increases with distance from the nearest anchor, so the model implicitly learns to modulate the strength of its refinement accordingly.
\cref{fig:mask_comparison} illustrates both effects, and quantitatively, removing the mask degrades PSNR by 1.3~dB (\cref{tab:mask_ablation}).

\paragraph{LoRA finetuning.}
We adapt the model using LoRA~\cite{hu2022lora} with rank $64$, updating less than 1\% of the total parameters while keeping the base model weights and the VAE frozen.
This minimal adaptation is sufficient because the task is narrower than general video generation: the model only needs to learn to condition on the degraded input, not to generate videos entirely from scratch.

\subsection{Training Data}
\label{sec:training_data}

We build training data of paired videos of degraded renderings and their clean ground-truth counterparts from DL3DV-10K~\cite{ling2024dl3dv}, which provides diverse videos with precomputed COLMAP~\cite{schonberger2016structure} reconstructions.
For each scene, we uniformly sample $k \in [3, 12]$ frames as training views and extract 61-frame trajectories that pass through at least two of them.
We then render each trajectory from four representation types to produce diverse degraded-clean pairs (\cref{fig:artifact_types}):

\begin{itemize}
    \item \textit{NeRF~\cite{mildenhall2020nerf}}: we run Nerfacto~\cite{nerfstudio} on the sparse training views, with characteristic blur and fog artifacts at novel viewpoints.
    \item \textit{3DGS~\cite{kerbl3Dgaussians}}: we initialize from a random point cloud and fit a model with gsplat~\cite{ye2025gsplat} for 7K iterations, deliberately underfitting so that novel viewpoints exhibit visible artifacts while training viewpoints render cleanly.
    \item \textit{Meshes}: we run MapAnything~\cite{keetha2026mapanything}, a feed-forward 3D reconstruction model, on the training views and fit a triangular mesh from the predicted depths. Frames at training viewpoints are replaced with the original images because depth is unreliable at sky regions and occlusion boundaries.
    \item \textit{Sparse point clouds}: we retain only COLMAP keypoints visible in the selected training views. Because these points exist only at detected keypoints and lack coverage in textureless regions (e.g., sky or walls), we replace frames at training viewpoints with the captured images to provide richer context.
\end{itemize}

\paragraph{Data efficiency.}
As the pretrained video model already encodes strong priors about videos, the training data only needs to teach it to condition on degraded input, a simpler task than learning video generation from scratch.
Therefore, LoRA finetuning requires remarkably little data: 20 paired videos already produce effective rendering cleanup, and scaling to 500 yields further improvements (\cref{tab:data_ablation}).
By comparison, prior methods require 80K--150K image pairs~\cite{wu2025difix3d+, liu20243dgs}.

\subsection{Geometry-Aware Preference Optimization}
\label{sec:geometry_dpo}

\begin{figure*}[t!]
    \centering
    \includegraphics[width=\linewidth]{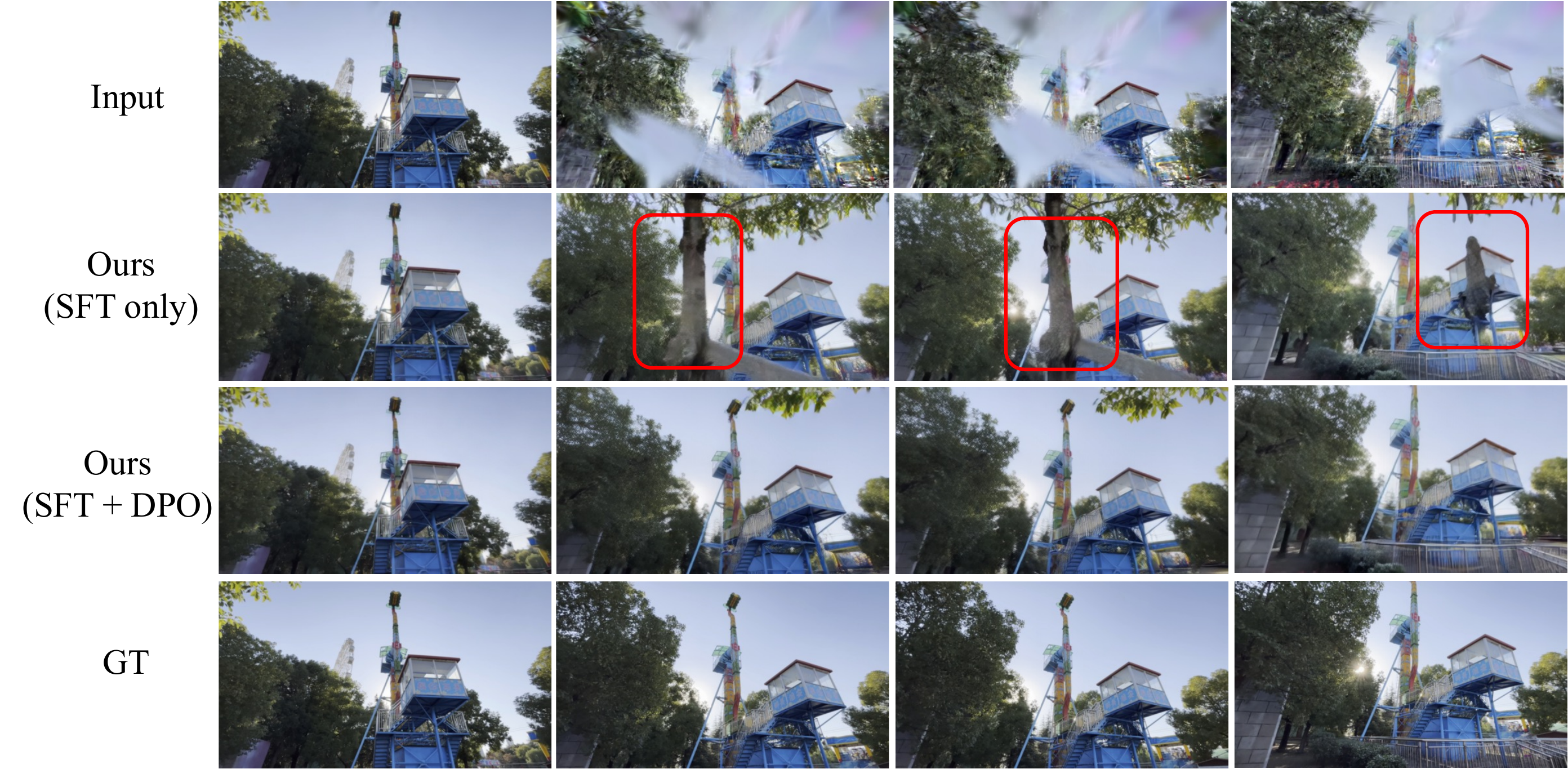}
    \vspace{-0.25in}
    \caption{\textbf{Why geometric-aware preference optimization helps.}
    The SFT-only model (row~2) hallucinates structures (red boxes) that shift across frames, causing SfM to recover wrong poses.
    After Flow-DPO (row~3), the model avoids such hallucinations and closely matches the ground truth (row~4), improving pose AUC@5\degree{} by 7.2\%.}
    \label{fig:dpo_comparison}
\end{figure*}

The flow matching loss (\cref{eq:flow_matching}) encourages per-frame visual quality but does not explicitly enforce multi-view geometric consistency.
In practice, the supervised fine-tuning (SFT) model sometimes hallucinates structures that look plausible in individual frames but are geometrically inconsistent across views.
\cref{fig:dpo_comparison} (row~2) shows a typical failure: the model generates a tree-like structure (red boxes) that shifts position and shape across frames.
When SfM (e.g., COLMAP~\cite{schonberger2016structure}) is run on such outputs, it tracks these inconsistent features and recovers incorrect camera poses, a signal that we can use to improve the model.

\paragraph{Pose accuracy as a reward.}
We quantify geometric consistency by measuring how well camera poses can be recovered from a generated video.
For each output, we run COLMAP~\cite{schonberger2016structure} with SuperPoint~\cite{detone18superpoint} features and LightGlue~\cite{lindenberger2023lightglue} matching to estimate their camera poses, and compare against the known ground-truth.
We report the Area Under the Curve at a 5\degree{} threshold (AUC@5\degree{}) over both relative rotation accuracy (RRA) and relative translation accuracy (RTA)~\cite{jin2021image}.
Geometrically consistent videos yield high AUC because SfM reliably recovers their poses; videos with hallucinated or inconsistent geometry produce low scores.

\paragraph{Constructing preference pairs.}
For a separate set of 1{,}000 DL3DV scenes, we generate five candidate outputs per scene using different random seeds and rank them by AUC@5\degree{}.
We construct preference pairs $(\mathbf{y}_w, \mathbf{y}_l)$ by pairing higher-ranked outputs against lower-ranked ones, retaining only pairs with an AUC gap of at least 0.2 to ensure a clear preference signal.

\paragraph{Flow-DPO training.}
We optimize the model using Flow-DPO~\cite{liu2025improving}, which adapts DPO~\cite{rafailov2023direct} to rectified flow models.
Let $\mathbf{v}^w = \boldsymbol{\epsilon}^w - \mathbf{z}_0^w$ and $\mathbf{v}^l = \boldsymbol{\epsilon}^l - \mathbf{z}_0^l$ denote the target velocity fields for the preferred and dispreferred samples.
The loss is:
\begin{equation}
    \mathcal{L}_{\mathrm{DPO}} = -\mathbb{E}\left[\log \sigma\!\left(-\frac{\beta}{2}\left(\Delta_w - \Delta_l\right)\right)\right],
    \label{eq:flow_dpo}
\end{equation}
where $\Delta_w = \|\mathbf{v}^w - \mathbf{v}_\theta(\mathbf{z}_t^w, t)\|^2 - \|\mathbf{v}^w - \mathbf{v}_{\mathrm{ref}}(\mathbf{z}_t^w, t)\|^2$ and $\Delta_l$ is defined analogously for the dispreferred sample, with $\mathbf{v}_{\mathrm{ref}}$ being the SFT (LoRA finetuned) checkpoint.
This loss steers the model toward outputs that SfM methods can reconstruct accurately.
Because the geometric prior is baked into the learned LoRA adapter during training, no pose estimation is needed at inference.

\subsection{Inference}
\label{sec:inference}

At inference time, the user provides a rendering video $\mathbf{x}$ and a mask $\mathbf{m}$ indicating which frames are clean.
FixAnything produces a cleaned video by sampling from Gaussian noise $\boldsymbol{\epsilon} \sim \mathcal{N}(\mathbf{0}, \mathbf{I})$ at $t = 1$ and integrating the learned velocity field $\mathbf{v}_\theta$ toward $t = 0$, with the rendering video and mask provided as conditioning via channel concatenation (\cref{eq:concat}).
The process follows the standard flow matching ODE, and we use 50 denoising steps by default.
For scenes with more frames than the model's temporal window, we split the video into overlapping chunks of 61 frames.
Notably, we observe that reducing the number of denoising steps to 5 still produces reasonable results with a 10$\times$ speedup (\cref{tab:steps_ablation}).

\section{Experiments}
\label{sec:experiments}

\subsection{Experimental Setup}
\label{sec:setup}

\paragraph{Implementation details.}
We finetune Wan2.1-I2V-14B~\cite{wan2025wan} using LoRA with rank 64 on 500 paired DL3DV-10K~\cite{ling2024dl3dv} videos at $288 \times 512$ resolution first, then upgrade to $480 \times 832$ with $T{=}61$ frames per video.
SFT training runs for 3000 iterations on a single H100 GPU.
Flow-DPO training uses the SFT checkpoint as reference and further optimizes the LoRA adapter for 2000 additional iterations.
At inference, we use 50 denoising steps by default and process videos in overlapping chunks of 61 frames.

\paragraph{Dataset and evaluation protocol.}
Following the test splits of 3DGS-Enhancer~\cite{liu20243dgs} and Xu et al.~\cite{xu2025novel}, we evaluate on 20 held-out scenes from DL3DV-10K dataset~\cite{ling2024dl3dv}, where we uniformly select 3, 6, or 9 frames as training views from each test scene.
The remaining frames (excluding training views) are sampled at every 8th frame to form the query set.
We report standard metrics PSNR, SSIM, and LPIPS~\cite{zhang2018unreasonable} to measure synthesized image quality, and AUC@5\degree{} of relative rotation accuracy (RRA) and relative translation accuracy (RTA) to measure geometric consistency of the generated videos.

\paragraph{Baselines.}
We compare against two groups of methods.
\textit{Sparse-view reconstruction methods} train a 3D representation  using few input views: 3DGS~\cite{kerbl3Dgaussians}, RegNeRF~\cite{Niemeyer2021Regnerf}, FreeNeRF~\cite{Yang2023FreeNeRF}, DNGaussian~\cite{li2024dngaussian}, and FSGS~\cite{zhu2024fsgs}.
\textit{Post-hoc enhancement methods} refine the output of an existing reconstruction (3DGS rendering): 3DGS-Enhancer~\cite{liu20243dgs}, Xu et al.~\cite{xu2025novel}, and Difix3D+~\cite{wu2025difix3d+}.
For FixAnything, we report results using four different input representations, all built from the same sparse training views following the protocol in \cref{sec:training_data}.
All four use the same single model, and only the rendering input changes.

\subsection{Comparison with Prior Methods}
\label{sec:comparison}

\begin{figure*}[t]
    \centering
    \includegraphics[width=\linewidth]{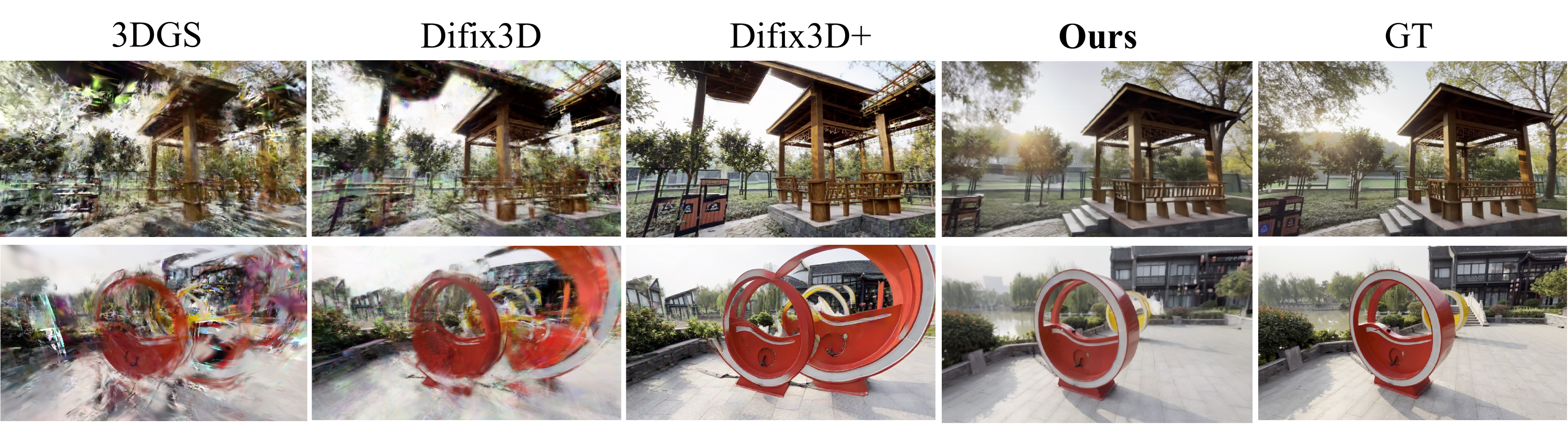}
    \vspace{-0.25in}
    \caption{\textbf{Qualitative comparison with Difix3D and Difix3D+~\cite{wu2025difix3d+}}.
    Difix3D enhances each frame independently using the nearest training view then distills results into 3DGS, introducing inconsistencies.
    Difix3D+ applies an additional enhancement step at render time, sharpening details but amplifying these inconsistencies.
    By processing the full video at once, FixAnything produces temporally coherent output.}
    \label{fig:difix_comparison}
\end{figure*}

\cref{tab:main_results} compares FixAnything against prior methods on DL3DV under 3-view, 6-view, and 9-view settings.
\begin{table*}[t]
\centering
\caption{\textbf{Quantitative comparison on DL3DV.}
FixAnything uses a \textit{single model} to clean up renderings from four different input representations (NeRF, 3DGS, mesh, sparse SfM points), achieving comparable quality across all four.
Notably, sparse SfM point renderings perform on par with or better than 3DGS and NeRF inputs, despite providing only scattered keypoints and the sole dense visual information comes from the few clean training views that the trajectory passes through.
We color each cell as \colorbox{first}{best} and \colorbox{second}{second best}.
($\dagger$~indicates numbers reported by~\cite{liu20243dgs,xu2025novel})}
\label{tab:main_results}
\vspace{-1mm}
\resizebox{\linewidth}{!}{%
\begin{tabular}{l ccc ccc ccc}
\toprule
& \multicolumn{3}{c}{3 Views} & \multicolumn{3}{c}{6 Views} & \multicolumn{3}{c}{9 Views} \\
\cmidrule(lr){2-4} \cmidrule(lr){5-7} \cmidrule(lr){8-10}
Method & PSNR$\uparrow$ & SSIM$\uparrow$ & LPIPS$\downarrow$ & PSNR$\uparrow$ & SSIM$\uparrow$ & LPIPS$\downarrow$ & PSNR$\uparrow$ & SSIM$\uparrow$ & LPIPS$\downarrow$ \\
\midrule
\multicolumn{10}{l}{\textit{Sparse-view reconstruction}} \\
3DGS~\cite{kerbl3Dgaussians}$^\dagger$ & 10.97 & 0.248 & 0.567 & 13.34 & 0.332 & 0.498 & 14.99 & 0.403 & 0.446 \\
RegNeRF~\cite{Niemeyer2021Regnerf}$^\dagger$ & 11.46 & 0.214 & 0.600 & 12.69 & 0.236 & 0.579 & 12.33 & 0.219 & 0.598 \\
FreeNeRF~\cite{Yang2023FreeNeRF}$^\dagger$ & 10.91 & 0.211 & 0.595 & 12.13 & 0.230 & 0.576 & 12.85 & 0.241 & 0.573 \\
DNGaussian~\cite{li2024dngaussian}$^\dagger$ & 11.10 & 0.273 & 0.579 & 12.67 & 0.329 & 0.547 & 13.44 & 0.365 & 0.539 \\
FSGS~\cite{zhu2024fsgs}$^\dagger$ & 12.22 & 0.296 & 0.535 & 13.73 & 0.429 & 0.540 & 15.52 & 0.468 & 0.416 \\
\midrule
\multicolumn{10}{l}{\textit{Post-hoc enhancement (3DGS rendering)}} \\
3DGS-Enhancer~\cite{liu20243dgs}$^\dagger$ & 14.33 & 0.424 & 0.464 & 16.94 & 0.565 & 0.356 & 18.50 & 0.630 & 0.305 \\
Xu et al.~\cite{xu2025novel}$^\dagger$ & 14.62 & \second 0.471 & 0.491 & 17.35 & 0.566 & 0.396 & 19.19 & 0.616 & 0.335 \\
Difix3D~\cite{wu2025difix3d+} & 12.85 & 0.392 & 0.557 & 14.84 & 0.445 & 0.462 & 16.76 & 0.520 & 0.399 \\
Difix3D+~\cite{wu2025difix3d+} & 12.37 & 0.363 & 0.512 & 14.41 & 0.424 & 0.400 & 16.39 & 0.498 & 0.330 \\
\midrule
\multicolumn{10}{l}{\textbf{\textit{FixAnything (Ours) -- single model}}} \\
\quad w/ NeRF rendering & 14.22 & 0.427 & 0.451 & 17.01 & 0.522 & 0.329 & 18.86 & 0.605 & 0.297 \\
\quad w/ 3DGS rendering & 15.18 & 0.452 & 0.408 & 17.65 & 0.561 & 0.289 & \second 19.76 & \second 0.632 & 0.269 \\
\quad w/ mesh rendering & \first 15.74 & \first 0.482 & \first 0.366
                         & \first 17.95 & \first 0.583 & \first 0.269
                         & \first 19.86 & \first 0.646 & \first 0.233 \\
\quad w/ sparse SfM points & \second 15.52 & 0.463 & \second 0.381 & \second 17.74 & \second 0.568 & \second 0.271 & 19.72 & 0.624 & \second 0.241 \\
\bottomrule
\end{tabular}
}
\end{table*}
Sparse-view reconstruction methods struggle in this setting: 3DGS produces visible floater artifacts, while regularization-based approaches (RegNeRF, FreeNeRF, DNGaussian) improve geometry but cannot complete content in under-observed regions.
Among post-hoc enhancement methods, 3DGS-Enhancer~\cite{liu20243dgs}, Xu~et~al.~\cite{xu2025novel}, and Difix3D+~\cite{wu2025difix3d+} demonstrate the value of generative priors, achieving substantial gains over all sparse-view baselines.
FixAnything achieves competitive or superior performance on the 3DGS input while being simpler to implement and easier to adapt to future video backbones.
\cref{fig:difix_comparison} shows a qualitative comparison with Difix3D and Difix3D+ which produce sharper individual images but struggle with cross-view consistency, whereas FixAnything produces temporally coherent output.

\begin{figure*}[t]
    \centering
    \includegraphics[width=\linewidth]{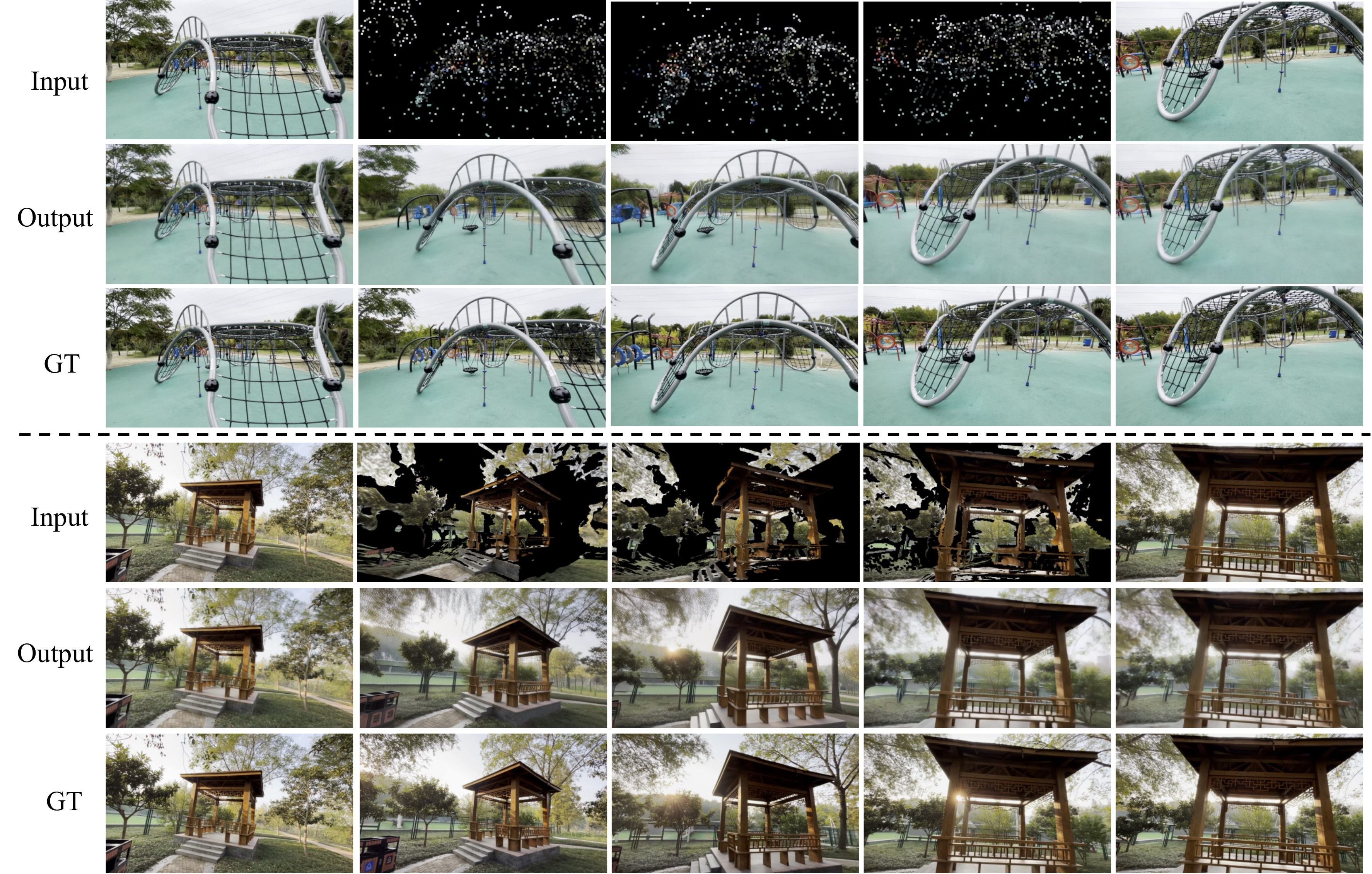}
    \vspace{-0.25in}
    \caption{\textbf{FixAnything generalizes across input representations.}
    The same model cleans up sparse COLMAP point clouds (top) and meshes (bottom), producing photorealistic output despite minimal visual input except for the clean first and last views.}
    \label{fig:colmap_result}
\end{figure*}

Notably, the same model also cleans up mesh renderings and sparse point clouds to comparable quality (\cref{tab:main_results}, bottom rows), despite these inputs providing far less visual information than 3DGS.
\cref{fig:colmap_result} shows qualitative results: even when the input consists of sparse COLMAP keypoints with clean
training views interspersed along the trajectory, the model produces dense, photorealistic output.
This supports the finding from \cref{sec:problem} that the rendering primarily serves as a structural scaffold while the video prior fills in the rest.


In supplementary materials, we further evaluate our model on MipNeRF-360~\cite{barron2022mipnerf360} and LLFF~\cite{mildenhall2019local}, where FixAnything (using 3DGS input) achieves comparable performance to SOTA methods~\cite{yu2024viewcrafter, xu2025novel} with a notable improvement in LPIPS, showing strong cross-dataset generalization.

\subsection{Effect of Geometry-Aware Preference Optimization}
\label{sec:dpo_results}
\vspace{-2mm}
\cref{tab:dpo_ablation} compares the SFT-only model against the full model after Flow-DPO training, with camera pose accuracy as a reward signal to construct preference pairs for Flow-DPO (\cref{sec:geometry_dpo}).
While image quality metrics (PSNR, SSIM, LPIPS) improve modestly, the primary gain is in geometric consistency: AUC@5\degree{} improves by 7.2\%, so COLMAP recovers more accurate camera poses from the DPO-refined outputs.
\cref{fig:dpo_comparison} illustrates why: the SFT-only model can produce frames that look clean but contain hallucinations (red boxes), causing SfM to fail.
After Flow-DPO training, these hallucinations are suppressed and the model produces geometrically coherent output closely matching the ground truth.
Crucially, this improvement comes at no additional inference cost as the geometric prior is baked into the model weights (LoRA adapter) during DPO training.

\begin{table}[t]
\begin{minipage}[t]{0.50\linewidth}
\centering
\caption{\textbf{Effect of Flow-DPO.} Finetuning with Flow-DPO improves geometric consistency at no extra inference cost.}
\label{tab:dpo_ablation}
\vspace{-6pt}
\begin{tabular}{l ccc c}
\toprule
Method & PSNR$\uparrow$ & SSIM$\uparrow$ & LPIPS$\downarrow$ & AUC@5\degree{}$\uparrow$ \\
\midrule
SFT only & 17.51 & 0.554 & 0.296 & 61.12 \\
+DPO & \textbf{17.65} & \textbf{0.561} & \textbf{0.289} & \textbf{68.32} \\
\bottomrule
\end{tabular}
\end{minipage}%
\hspace{8pt}%
\begin{minipage}[t]{0.46\linewidth}
\centering
\caption{\textbf{Mask-aware conditioning.} Without the mask, quality degrades due to hallucination over clean frames.}
\label{tab:mask_ablation}
\vspace{-6pt}
\begin{tabular}{l ccc}
\toprule
Variant & PSNR$\uparrow$ & SSIM$\uparrow$ & LPIPS$\downarrow$ \\
\midrule
No mask & 16.37 & 0.525 & 0.311 \\
With mask & \textbf{17.65} & \textbf{0.561} & \textbf{0.289} \\
\bottomrule
\end{tabular}
\end{minipage}
\end{table}

\subsection{Ablation Studies}
\label{sec:ablations}

We ablate the remaining design choices on DL3DV-10K with 6 training views.
\paragraph{Mask-aware conditioning.}
\cref{tab:mask_ablation} compares a variant with all mask entries set to 1 (no trust/fix distinction) against the full model.
Providing the mask improves PSNR by 1.3~dB: without it, the model must infer degradation severity from the visual signal alone, which is ambiguous since some clean frames resemble mildly degraded ones, causing hallucination over content that should be preserved.

\paragraph{Data efficiency and inference speed.}
\cref{tab:data_ablation} shows that 20 paired videos already produce effective cleanup, with diminishing returns beyond 100, showing that the pretrained model already captures most necessary priors and the paired data primarily teaches the conditioning mechanism.
\cref{tab:steps_ablation} shows that reducing denoising steps from 50 to 5 yields comparable quality across all metrics while providing a $10\times$ speedup, generating a 61-frame clip at $480{\times}832$ resolution in 31 seconds on a single H100.
This opens the door to further acceleration through distillation~\cite{yin2024one}, potentially enabling real-time rendering cleanup.

\begin{table}[t]
\begin{minipage}[t]{0.48\linewidth}
\centering
\caption{\textbf{Training data volume.}
As few as 20 paired training videos produce effective 3DGS rendering cleanup, with diminishing returns beyond 100 videos.}
\label{tab:data_ablation}
\vspace{-6pt}
\begin{tabular}{c ccc}
\toprule
Training vids. & PSNR$\uparrow$ & SSIM$\uparrow$ & LPIPS$\downarrow$ \\
\midrule
20  & 16.70 & 0.531 & 0.309 \\
50  & 17.20 & 0.548 & 0.297 \\
100 & 17.45 & 0.556 & 0.292 \\
500 & 17.65 & 0.561 & 0.289 \\
\bottomrule
\end{tabular}
\end{minipage}\hfill
\begin{minipage}[t]{0.48\linewidth}
\centering
\caption{\textbf{Inference denoising steps vs.\ runtime.}
Quality remains comparable even at 5 denoising steps, enabling a significant speedup.}
\label{tab:steps_ablation}
\vspace{-6pt}
\begin{tabular}{c ccc c}
\toprule
Steps & PSNR$\uparrow$ & SSIM$\uparrow$ & LPIPS$\downarrow$ & Time (s) \\
\midrule
5  & 18.02 & 0.574 & 0.313  & 31 \\
10 & 17.91 & 0.570 & 0.296 & 62 \\
25 & 17.75 & 0.564 & 0.289 & 155 \\
50 & 17.65 & 0.561 & 0.289 & 309 \\
\bottomrule
\end{tabular}
\end{minipage}
\end{table}

\vspace{-0.1in}
\section{Discussion: Hallucination and Uncertainty}
\label{sec:discussion}
\vspace{-0.05in}

A natural concern with generative cleanup is \emph{hallucination}.
We define \emph{hallucination} as content the model must invent when it is not seen in the input views.
However, we emphasize that hallucination is not inherently a failure: it is what makes generative cleanup useful by plausibly filling in unobserved regions, and it becomes a failure only when it contradicts the existing observations.

\begin{figure}[t]
  \centering
  \includegraphics[width=0.9\linewidth]{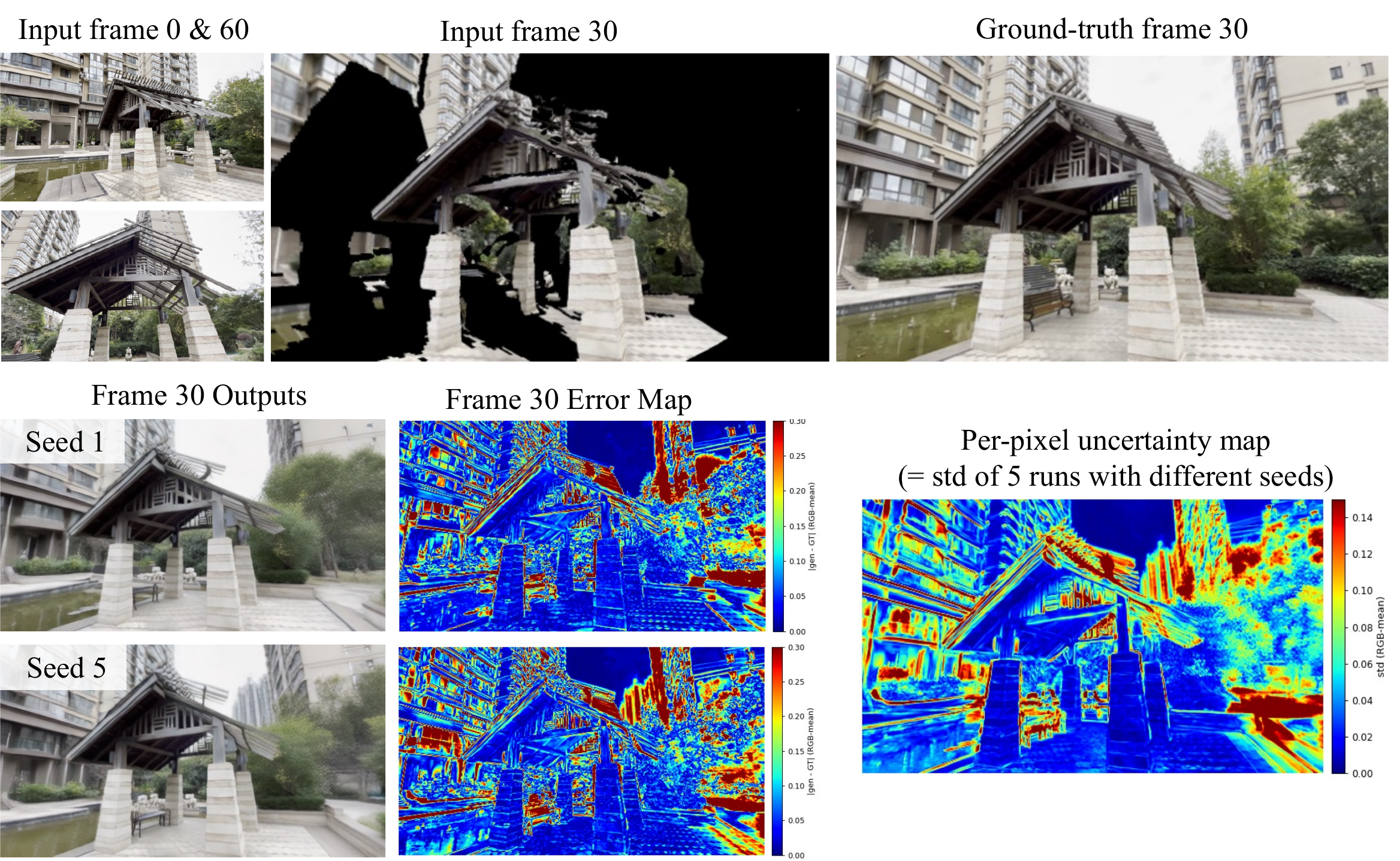}
  \vspace{-0.1in}
  \caption{\textbf{Uncertainty estimation.} Uncertainty maps of mesh rendering cleanup correlates well with reconstruction error and with unobserved (black) input regions.}
  \label{fig:unc}
\end{figure}

This raises a practical question: can we tell where the model is hallucinating?
As a preliminary analysis, we run inference $N{=}5$ times with different random seeds and use the per-pixel standard deviation as an uncertainty estimate (\cref{fig:unc}; black pixels denote occluded or unobserved regions).
Sky- and ground-like regions show \emph{low} uncertainty, where the model confidently propagates the input texture, while regions with multiple plausible completions (e.g., buildings) show \emph{high} uncertainty.
This uncertainty measurement also correlates well with regions of high reconstruction error: on DL3DV (6 views), the mean PSNR over the most-confident 25\% of pixels is 25.7\,dB versus 14.4\,dB over the least-confident 25\%.
This points to a simple, training-free way to quantify uncertainty, a promising step toward making generative cleanup more reliable.

\vspace{-0.1in}

\section{Conclusion}
\label{sec:conclusion}
We present a single model for removing artifacts across multiple 3D representations, including 3DGS, NeRF, meshes, and point clouds, by finetuning a pretrained video diffusion model on a modest set of paired videos. Our results suggest that many representations can be cleaned up equally well, raising questions about common reconstruction pipelines. Given $N$ input images, existing workflows typically estimate camera poses with COLMAP, learn a representation such as NeRF or 3DGS from the posed views, and finally refine rendered views with a generative model. Our experiments indicate that comparable results can be obtained by instead rendering sparse COLMAP point clouds (optionally meshed via multi-view stereo) and applying generative cleanup directly.

More broadly, our observations suggest that the main difficulty in novel-view rendering arises in regions that are weakly observed or unobserved. In these areas, classical reconstruction methods struggle because the problem becomes one of plausible completion rather than geometric inference. This points to an alternative strategy: reconstruct reliable geometry where observations exist, and use generative models to infer missing content where they do not. Future work could explore feeding such generative predictions back into reconstruction pipelines to produce more complete and consistent 3D scene models.

\small \noindent \textbf{Acknowledgments} 
We thank Shubham Tulsiani, Nikhil Keetha, Sriram Narayanan, Anurag Ghosh, and other members of Deva’s and Srinivas’ groups at CMU for their valuable feedback and suggestions at various stages of this project.
This work used Bridges-2~\cite{bridges2} at Pittsburgh Supercomputing Center through allocation \texttt{cis240119p} from the Advanced Cyberinfrastructure Coordination Ecosystem: Services \& Support (ACCESS) program, which is supported by National Science Foundation grants \#2138259, \#2138286, \#2138307, \#2137603, and \#2138296.
This work was supported by Intelligence Advanced Research Projects Activity (IARPA) via Department of Interior/Interior Business Center (DOI/IBC) contract number 140D0423C0074. The U.S. Government is authorized to reproduce and distribute reprints for Governmental purposes notwithstanding any copyright annotation thereon. Disclaimer: The views and conclusions contained herein are those of the authors and should not be interpreted as necessarily representing the official policies or endorsements, either expressed or implied, of IARPA, DOI/IBC, or the U.S. Government.

%
%
\bibliographystyle{splncs04}
\bibliography{main}

\end{document}